# Reinforcement Learning-Based Cooperative P2P Power Trading between DC Nanogrid Clusters with Wind and PV Energy Resources

Sangkeum Lee[a], Sarvar Hussain Nengroo[b], Hojun Jin[b], Taewook Heo[a], Yoonmee Doh[a], Chungho Lee[a], Dongsoo Har[b]

[a]Environment ICT Research Section, Electronics and Telecommunications Research Institute (ETRI), Daejeon, South Korea (email: sangkeum@etri.re.kr, htw398@etri.re.kr, ydoh@etri.re.kr, leech@etri.re.kr)

[b]Cho Chun Shik Graduate School of Mobility, Korea Advanced Institute of Science and Technology (KAIST), Daejeon, South Korea (email: sarvar@kaist.ac.kr, hjjin1995@kaist.ac.kr, dshar@kaist.ac.kr)

*Abstract*— In replacing fossil fuels with renewable energy resources for carbon neutrality, the unbalanced resource production of intermittent wind and photovoltaic (PV) power is a critical issue for peer-to-peer (P2P) power trading. To address this issue, a reinforcement learning (RL) technique is introduced in this paper. For RL, a graph convolutional network (GCN) and a bi-directional long short-term memory (Bi-LSTM) network are jointly applied to P2P power trading between nanogrid clusters, based on cooperative game theory. The flexible and reliable DC nanogrid is suitable for integrating renewable energy for a distribution system. Each local nanogrid cluster takes the position of prosumer, focusing on power production and consumption simultaneously. For the power management of nanogrid cluster, multi-objective optimization is applied to each local nanogrid cluster with the Internet of Things (IoT) technology. Charging/discharging of an electric vehicle (EV) is executed considering the intermittent characteristics of wind and PV power production. RL algorithms, such as GCN- convolutional neural network (CNN) layers for deep Q-learning network (DQN), GCN-LSTM layers for deep recurrent Q-learning network (DRQN), GCN-Bi-LSTM layers for DRQN, and GCN-Bi-LSTM layers for proximal policy optimization (PPO), are used for simulations. Consequently, the cooperative P2P power trading system maximizes the profit by considering the time of use (ToU) tariff-based electricity cost and the system marginal price (SMP), and minimizes the amount of grid power consumption. Power management of nanogrid clusters with P2P power trading is simulated on a distribution test feeder in real time, and the proposed GCN-Bi-LSTM-PPO technique achieving the lowest electricity cost among the RL algorithms used for comparison reduces the electricity cost by 36.7%, averaging over nanogrid clusters.

# I. INTRODUCTION

The widespread use of distributed energy resources (DERs) has significantly altered how energy is generated, transported, and used along the energy pipeline. A more decentralized and open electrical network is made possible with increased number of prosumers—individuals who produce and consume energy simultaneously. As a result of this context, new opportunities and difficulties for power systems have emerged. Peer-to-peer (P2P) power trading is a novel paradigm of distribution systems with a utility grid (UT) related to carbon neutrality and renewable energy generation [1]. P2P power trading has become a viable alternative for prosumers looking to actively participate in the energy market. Moreover, P2P trading gives end users more flexibility, increases possibilities to use clean energy, and aids in the transition to a low-carbon energy system. In addition to this, the other participants in the power market can also profit by lowering the peak electricity demand, lowering operating and maintenance expenses, and enhancing the dependability of the electrical system. In [2], an electricity pricing model was applied to P2P power trading to improve the community and reliability of the distribution network. Energy market design was investigated to implement P2P power trading for the integration of intermittent renewable energy sources (RESs) and battery flexibility [3]. A study in [4] presented a P2P market framework to gradually increase the number of renewable energy exchanges and settlements in microgrids.

To effectively manage the adoption of DERs, local distribution marketplaces have been proposed. A DC nanogrid was integrated with DER systems, using photovoltaic (PV) systems, energy storage systems (ESSs), and electric vehicles (EVs) [5]. In [6], the setup of the nanogrids for simulation was presented. The operation control of electric appliances was determined by the occupant location, and behavior pattern set by the hidden Markov model and the probability model of electric appliance usage. In [2], an electricity pricing model was applied to P2P power trading to improve the reliability of the distribution network. Energy market design was investigated to implement P2P power trading for the integration of intermittent RESs and battery flexibility [3]. A study in [4] presented a P2P market framework to gradually increase the number of renewable energy exchanges and settlements in microgrids.

The energy production from RESs like PV panel and wind turbine is sporadic in nature [7]. The diversity pattern of local wind power generation does not guarantee a stable supply in the electricity network [8]. Wind power producers provide energy to a demand response (DR) aggregator through P2P power trading and the successful supplier gets the maximum profit. A P2P local electricity market model was presented considering joint energy trading and uncertainty in [9]. To balance the uncertainty of PV power production locally in the market, prediction of time-flexible demand is effective. The test results showed that a PV forecast error is balanced locally with the proposed joint market. Demand-side response [10], time of use (ToU) tariff, and direct load control are applied for the network reinforcement that needs large investment and much lead time [2].

In the nanogrid clusters, the charging/discharging of EV plays the role of both prosumer/consumer, which improves the reliability and stability of the power system. The vehicle-to-grid (V2G) technology was utilized to stabilize the electricity network with the charging/discharging scheduling of EVs [11]. In [12], decentralized power system was established through P2P power trading, and $CO_2$ emission was significantly reduced by combining a PV system and EV penetration for carbon neutrality. A study in [13] presented an operational framework of P2P power trading for EVs in a charging station and a business association with a PV system to improve the benefits of all parties joining P2P power trading.

Individual use of RESs lacks reliability in sustainable operation [14]. Hybrid renewable systems are more reliable for P2P power trading. A hybrid wind-PV power system is considered as a secondary source of power supply in the grid-connected environment. Furthermore, the hybrid wind-PV-battery system can perform better in reducing the cost and improved reliability [15]. The hybrid wind-PV system is considered as eco-friendly economic resource in the distributed power system. In [16], the non-smooth and non-convex operation costs and emissions of power units in a complex system were minimized through the power management subjected to prohibited operating zones and multi-fuel options.

The architecture of the P2P power trading system is composed of a power network, an information network, and a business network. Distributed energy storage, dispatchable load, EV, and general loads are integrated to joint points in the power network, which is made up of networked microgrids located in various geographical areas. Internal architecture of each microgrid is unique, and the microgrids can work alone or in tandem. Information network is established with communication instruments, protocols, and data transfer. Thanks to smart meters deployed in each microgrid, the real-time energy consumption is recorded in the distribution network database. The business network is a P2P network with a blockchain that allows the development of multiple business models that determine the energy transaction with peers and the grid. The advancement of RESs has paved the way for distributed P2P power trading in different places like homes and buildings [17]. P2P power trading research has focused on the stability of the distribution system as the best bidding technique. Several types of P2P power trading architecture with DER systems, such as local, centralized, decentralized, distributed, or hierarchical, were utilized for more schedulable and reliable power system [18]. In local P2P power trading, each peer determined which peers would carry out the P2P power trading, depending on trading objectives [19]. Decentralized P2P power trading that can prevent network problems such as overvoltage might be robust and scalable for power transactions. In centralized P2P power trading, a single controller was used for an overall power network composed of communication links. The energy allocation algorithm was applied to a centralized P2P power trading market to achieve market equilibrium, maximizing global welfare [20]. In [21], authors proposed a P2P energy financial space to encourage prosumers to establish merged power plants. The authors in [22] devised an energy exchanging framework that takes into account the economic benefits while improving the reliability and security of the power system. Some studies have dealt with P2P energy transactions involving prosumers [23, 24]. In distributed P2P power trading, peers communicate with local peers and implement bidding through a distributed management system using an integrated communication network and advanced control strategies [25]. A hierarchical P2P architecture was presented for a tree-shaped feedback control system in [26].

Microgrid operators are permitted to get involved in P2P power trading in the local distribution network under the law of energy market for distribution network operator. In [27], a parallel bidding framework was created to facilitate energy trade across microgrids based on a three-layer distribution network architecture and the decentralized properties of a blockchain. With the advent of blockchains, it is possible to safely automate the process of P2P power trading. Blockchain is a decentralized and open ledger that maintains affairs in an efficient, permanent, and verifiable manner [28]. A new blockchain platform framework of P2P power trading to connect producers' resources and consumers' needs was examined in [29]. To increase the vitality of the market and complete the process, the framework of the proposed blockchain platform divided into a three-dimensional platform provided users with a visual interface as well as an equipment inspection system. An amalgamated energy blockchain was proposed for safe P2P power trading, and a credit-based payment mechanism was created to manage energy traffic system while allowing quick payment [30, 31]. A recent study in [32] suggested validated key management methods in smart grids and investigated them for security risks, and a safe and substantiated keyless method based on blockchain was presented. A game-theoretic method for the demand-side management model was devised in [33], including storage components and blockchain technology used for regulation and reliable operation.

The Internet of Things (IoT) was applied for power management to monitor and obtain sensor data in real time through the internet [34]. IoT sensors collect physical data such as voltage and current in the cloud or cluster of power systems, and the collected data is sent to the server to be saved or used for system management [35]. It also extends the advantages of continuous internet connection for interconnected systems to implement data sharing, data acquisition, and remote control of connected electric appliances, EVs, and Heater, Ventilation fan, and Air-Conditioner (HVAC). In IoT applications, two processes that obtain process control and monitoring of data are often used. To monitor data, hypertext transfer protocol and message queueing telemetry transport protocol can be used as communication protocols.

The basis of distributed P2P power trading has developed as a next generation energy management tool for the smart grid, and it empowers each prosumer of the organization to take interest in exchanging energy with other prosumers and the UT. P2P power exchanging is being considered as a likely apparatus to advance the utilization of DERs, and the fundamental goal of P2P power sharing is to break the unified foundation of the power grid by permitting the transmission and supply of energy between different prosumers [36]. This allows interested customers to purchase eco-friendly power at a lower cost from a peer who has extra renewable energy, which decreases their reliance on the grid or a central provider. In [37], a day-ahead optimization procedure managed by an independent principal unit was proposed for decreased expense of energy exchanging.

Reinforcement learning (RL) is a subset of machine learning techniques and makes a sequence of decisions that enables an agent to learn how to achieve a goal in uncertain and complex environment [38]. In recent years, with the growing interest in artificial intelligence, the RL has been explored to solve the decision-making problems in smart grids. In fact, the RL is being utilized in many areas such as the robot control, transportation, power management, and operational research [39]. Unlike other commonly used approaches such as model predictive control and optimal methods, RL approaches are able to achieve robust real-time performances. To minimize the electricity cost of a microgrid that satisfies the power balance and the limit of generation units, dynamic hierarchical RL was proposed to implement optimal policy exploration [40]. In [41], a fully-automated algorithm based on the RL was used for an energy management system that makes optimal decisions for consumers. A model-free batch RL algorithm was used to schedule a cluster of domestic electric water heaters using the thermal storage of water tanks [42]. In [43], an RL-based energy trading framework was described to maximize individual average revenue in each microgrid by individually and randomly choosing a trading method. RL for energy sharing systems has also been proposed to maximize the energy exchange with the consideration of user model [44]. Deep Q-learning for a local energy trading algorithm was proposed with a prosumers' energy trading behavior model [45]. In [46], an RL algorithm was used to determine the action needed at a specific moment to maximize the profit. Prosumers were modeled to optimize their trading by employing a wind turbine and an ESS, and they were given the option of charging or discharging the ESS, as well as buying or selling energy in the local energy market. The investigation in [47] came up with an improvement model to assess P2P multiple energy exchanging among private and business prosumers, considering joint demand-side management. Both power and heat were exchanged, and the trade costs were advanced by a Nash-type method with ensured advantage allotment between two prosumers.

Different strategies and control techniques have been studied to improve P2P power trading performance. The authors in [23] proposed an energy management technique that enables prosumers to trade energy in the P2P market, considering trading preferences. With the implementation of price oriented optimization and predictive control, system needs were automatically adjusted in response to changing electricity cost over time. To optimize the EV/BESS charging schedule of a single end user, a DR based on a mixed-integer linear programming model was proposed [48]. This type of adaptable trading decision-making in connection to market price may be able to change their load profiles and reduce their overall energy costs. Although the energy management trading approaches can effectively increase the consumption rate of local renewable energy, the trade-off trading flexibility will undoubtedly lower the financial incentives for market participants. Recently, several bidding tactics making use of game- and auction-theoretical techniques have been studied to address these shortages. The utility of both suppliers and customers was taken into account in a game-theoretic approach concerned with the Nash equilibrium (NE) to maximize ToU pricing [49]. However, the ToU price simply represented the total market clearing price and did not account for real time competitive pricing interactions, making it difficult to directly apply to the scenario of P2P power trading in nanogrids.

In this paper, we propose a P2P power trading method by using RL algorithms. In the P2P power trading system, the DR of the UT and the system marginal price (SMP) of the RESs are considered at a near-maximum profit. The nanogrid cluster is integrated with secondary sources (wind and PV power generation) providing power to other nanogrid clusters through the distribution system. P2P power trading and a hybrid wind-PV system improve the inherent intermittency of renewable energy generation. An RL algorithm is utilized to determine selling/buying operations to/from the UT/RESs in each nanogrid cluster to achieve the energy efficient and economical P2P power trading. In each nanogrid cluster, the maximum load is minimized to maintain a reliable electricity network and the utilization of RESs is maximized to decrease grid power consumption through multi-objective optimization. To maximize the benefit of the DC nanogrid and establish a plan for its optimal operation, the EV is charged using surplus power during time periods of high renewable power generation, and the EV is discharged during time periods of low renewable power generation. The P2P power trading of an energy prosumer model is integrated with the V2G system. The building operator is a demanding customer who receives electricity from the Korea Electric Power Corporation (KEPCO), and at the same time, it is possible to trade discharging power of the EV by signing an agreement with the user for charging equipment. The main contribution and novelty of this study on the proposed P2P power trading method for nanogrid clusters are as follows

- A P2P power trading method using RL with cooperative game theory is proposed for DC nanogrid clusters. The proposed method increases the flexibility of connecting UT with RESs and the reliability for real-time power management between nanogrid clusters in an unbalanced environment involving different types and amounts of resource production. RL algorithms can save electricity cost compared with the rule-based P2P trading, which is considered as the baseline P2P trading, and also works well in conjunction with multi-objective optimization for nanogrid cluster.

- In nanogrid clusters, RL algorithms, such as deep Q-learning network (DQN), deep recurrent Q-learning network (DRQN), bi-directional long short-term memory (Bi-LSTM) network and deep recurrent Q-learning network (Bi-DRQN), and proximal policy optimization (PPO) are proposed for a prosumer model in a P2P power trading system. The RL algorithms are used in place of P2P trading that is a part of power management. For enhanced performance of P2P power trading, a graph convolutional network (GCN) and a Bi-LSTM network are also integrated with RL algorithms. The P2P power trading method is examined in two cases. In the case 1(*DQN, DRQN, Bi_DRQN, PPO*), RL-based cooperative P2P power trading, e.g., the selling/buying operation of each cluster, is determined considering only the SMP, and in the case 2 (*N_DQN, N_DRQN, N_Bi-DRQN, N_PPO*), proposed RL-based cooperative P2P power trading, the selling/buying operation of each cluster is determined according to real-time fluctuations of the DR related to carbon neutrality and SMP, which achieves more flexible power management. Each layer of RL-based cooperative P2P power trading is based on the GCN-convolutional neural network (CNN) layers (*DQN, N_DQN)*, GCN-LSTM layers (*DRQN, N_DRQN)* and GCN-Bi-LSTM layers (*Bi_DRQN, N_Bi_DRQN, PPO, N_PPO)*.

- The hybrid wind-PV system for P2P power trading is concerned with yearly power of the UT according to seasonal patterns (spring/summer/autumn/winter). For maximizing the profit from P2P power trading, RESs as well as the UT is considered in RL-based cooperative P2P power trading.

- For real-time P2P power trading between nanogrid clusters, a specific procedure for P2P power trading, which is composed of six stages, information collection, registration, routing, scheduling, transmission, and settlement is introduced. The nanogrid cluster system consists of two domains, cyber and physical domains, and in other aspect, it is divided into three layers, application layer, router layer, and power layer.

- The proposed EV charging/discharging strategy in conjunction with P2P power trading provides electricity cost reduction depending on the trend of wind and PV power generation in real time.

The rest of this paper is organized as follows. In Section II, the nanogrid cluster architecture is explained. Section III provides details of RL algorithms for cooperative P2P power trading. Section IV presents the simulation results of the proposed P2P power trading and in Section V, concluding remarks are presented.

## II. MODELING OF THE NANOGRID OPERATION

*A. Nanogrid energy cloud system*

Each nanogrid consists of electric loads placed in four rooms, UT, RESs(wind turbine and a rooftop PV system), and electric loads can be scheduled or unscheduled. A list of schedulable(or flexible) and non-schedulable(or non-flexible) loads, along with power ratings and room indices, is shown in Table 1 [6]. In the nanogrid, utilization of flexible loads can be delayed to the next time interval to reduce peak power consumption. However, the power needs to be provided immediately to non- flexible loads when required. To define transition probabilities characterizing random movement of resident between rooms, a Markov chain (MC) model is adopted. It is assumed that one user(resident) is located in each nanogrid and the user moves to another room or stays in the same room at the end of each time period, depending on the transition probabilities. Occupant behavior of using electric appliances in rooms is also determined by the emission probability of the MC, based on the Korean Time Use Survey (KTUS) data [50]. The probabilistic usage of electric loads, as established by the data from the KTUS, governs how residents use electric appliances in their rooms. As shown in Table 1, the HVAC are placed in all 4 rooms, being denoted by HVAC(1,2,3,4). For details of the MC used for the nanogrid, readers are referred to [6].

TABLE 1. List of non-flexible and flexible electric appliances with their power ratings and room indices.

| CATEGORY | ELECTRIC APPLIANCES (ROOM LOCATION) | POWER | CATEGORY | ELECTRIC APPLIANCES (ROOM LOCATION) | POWER |
|---|---|---|---|---|---|
| NON-FLEXIBLE | AIR-CONDITIONER IN HVAC (1,2,3,4) | 1.21kW | FLEXIBLE | WASHING MACHINE (3) | 242W |
| NON-FLEXIBLE | ELECTRIC FAN IN HVAC (1,2,3,4) | 60W | FLEXIBLE | VACUUM CLEANER (2) | 1.07kW |
| NON-FLEXIBLE | HEATER IN HVAC (1,2,3,4) | 1.16kW | FLEXIBLE | IRON (2) | 1.23kW |
| NON-FLEXIBLE | COMPUTER (4) | 255W | FLEXIBLE | MICROWAVE OVEN (3) | 1.04kW |
| NON-FLEXIBLE | TV (2) | 130W | FLEXIBLE | RICE COOKER (3) | 1.03kW |
| FLEXIBLE | AUDIO (1) | 50W | FLEXIBLE | HAIR DRYER (4) | 1kW |

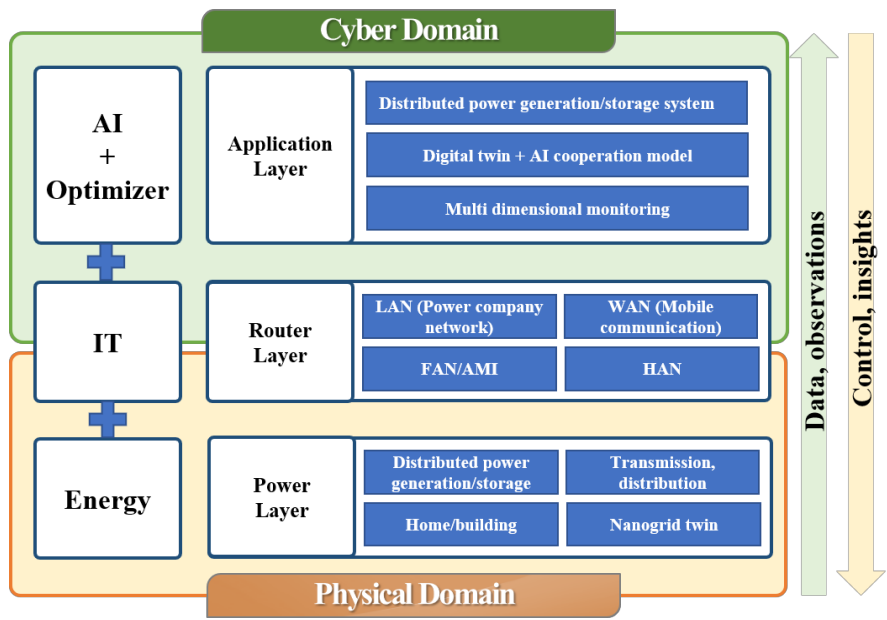

Fig. 1. P2P power trading system architecture of the nanogrid cluster.

The P2P power trading occurs between clusters, each consisting of nanogrids. The P2P power trading between nanogrid clusters regulates the power distribution system in real time and has high reliability and low power loss, which is advantageous from the viewpoint of each constituent DC nanogrid. The overall system architecture is composed of six nanogrid clusters, and each one is composed of three nanogrids. All clusters are electrically and physically connected to each other for P2P power trading. One wind turbine, represented by three wind generators, and three rooftop PV systems are operated as a cluster with their associated electric appliances. Each nanogrid cluster has a smart meter installed to track, record, and send data of load demand and renewable energy generation. It also uses a smart contract protocol for P2P power trading to connect with other clusters, and P2P power trading can take advantage of data collected by the smart meters.

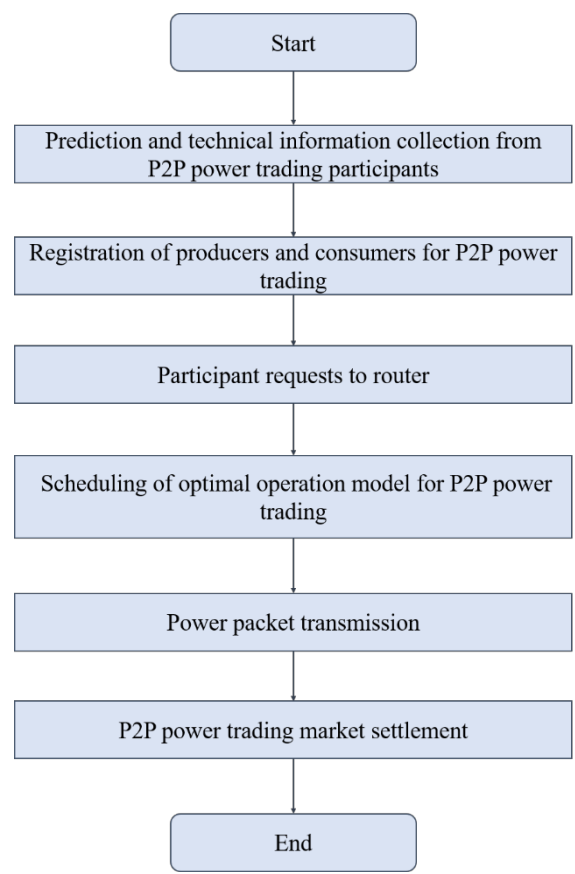

Fig. 2. P2P power trading procedure for nanogrid clusters.

The nanogrid cluster architecture shown in Fig. 1 is primarily separated into cyber and physical domains, and has three layers (application, router, and power). The physical domain includes different energy sources. All four components of energy flow, e.g., distributed power generation/storage, transmission and distribution, homes and buildings, and nanogrids, are controlled by the physical infrastructure. Distributed generation resources contain RESs like wind and PV energy. Transformers and substations are examples of energy transmission equipment that are part of the energy transmission system. Physical domain in the energy consumption area consists of different types of loads, including industrial power loads, household appliances, EVs, building heating systems, and so on. Cyber domain monitors and controls the physical infrastructure components. Combining AI with optimization technologies can be used for multi-dimensional monitoring, digital twins, and cooperative AI models. The communication ports of the router layer assist multiple energy stakeholders in making decisions. For instance, IoT infrastructure can gather data on the energy generation side and track the condition of the energy network via wide-area network on the energy transmission side. Moreover, it can gather information such as appliance level energy consumption data of buildings on the energy consumption side and can track the SOC of the EV on the energy storage side. A common application of the IoT paradigm in the energy industry has been the widespread deployment of advanced metering interfaces and other network sensing devices in various types of networks, such as home area networks. In the router layer, each nanogrid cluster shares data (renewable power generation and load demand) using HTTP and MQTT. P2P power trading between the clusters is established based on the data collected by the smart meter in the router layer. In the power layer, power is generated by a distributed power generation system, imported from the UT system, and purchased from other clusters through a P2P power trading system. Consequently, it is possible to implement a P2P power trading system with a three-layer system composed of the application layer, router layer, and power layer. Depending on the load demand associated with wind and PV power generation, each cluster can be a consumer or/and a producer in the inter-cluster power trading.

The cost of electricity is determined by DR programs, such as the ToU tariff from the UT and the system marginal price (SMP) from RESs. Customers who use less electricity during peak demand are given incentives through the DR program. Thus, the DR program makes it possible to significantly lower the cost of electricity and increase the dependability of the power systems. The KEPCO sets the electricity pricing in accordance with the DR program in the Republic of Korea. Fig. 4(a) shows the electricity rate of DR program (ToU tariff) and average monthly SMP. As shown in Fig. 4(a), the rate of electricity is $0.05/kWh over the time interval 23:00-09:00, $0.1/kWh over the time intervals 09:00-10:00, 12:00-13:00, and 17:00–23:00, and $0.18/kWh over the time intervals 10:00-12:00 and 13:00-17:00. The power produced by RESs can be exported to other nanogrid clusters in the P2P power trading system at the SMP. The SMP is utilized as the buying and selling price of the electricity produced by RESs and relates to the cost of the most expensive producing unit in the price setting schedule. In the Republic of Korea, prosumers are referred to as those who install renewable power generation systems with a capacity of less than 1MW. In addition, P2P power trading with electricity vendors is allowed, with no requirement for centralized communication.

*B. P2P power trading procedure for nanogrid cluster*

The P2P power trading system is operated with the power packet transfer architecture. The utilization of hybrid RES (wind and PV system) for P2P power trading is optimized by cooperative game theory between nanogrid clusters which can be assigned to a producer or a consumer. The procedure for P2P power trading is composed of six stages: information collection, registration, routing, scheduling, transmission, and settlement. In the information collection stage, participants in the P2P power trading provide predictive and technical information. In the registration step of P2P power trading, each nanogrid cluster is registered as a producer or a consumer depending on the information gathered by the smart meter. Through the routing step, power packets from the producers are sent to consumers connected with neighboring routers. In the scheduling step, the optimal P2P power trading model is established and the intermediate router transmits the power packet of the consumer in the transmission stage. With a P2P power trading controller, the P2P power exchange is implemented in the settlement stage.

We used game theory for the proposed P2P power trading system, which uses mathematical models to analyze strategic interactions by rational decision-makers. It can be mainly applied to non-cooperative/cooperative game applications [51-53]. The non-cooperative game theory investigates the crucial choices made by competing players as a result of their interactions, in which each player chooses its strategy for improving its performance or minimizing its losses (costs). Non-cooperative games can be solved using a variety of methods, including the well-known Nash equilibrium [52]. The authors in [54] presented a non-cooperative game that implements P2P power trading among microgrids. However, the cooperative game theory is analyzed with the framework to predict which coalitions will be formed, the actions to take, and the overall benefits. It is implemented in a high-level approach as it only describes the architecture, strategies, and advantages of formed coalitions. In [55], a cooperative power dispatching algorithm was presented for microgrids to manage power networks economically and to satisfy stochastic load demand within the UT. For long-term power management, the cooperative game framework for all clusters is more significant to increase energy efficiency and to facilitate the power consumption of DERs.

The P2P power trading system integrates a renewable energy generation system with an EV considering the intermittent issues. The role of peers is assigned to the participants of the P2P power trading system (e.g., producers or consumers). It focuses on the interaction between power supply and load demand, taking into account the various aspects of the community. The power each producer provides for P2P power trading is proportional to the amount of power generated by each producer. In addition, the power each consumer demands through P2P power trading is proportional to the amount of power required by each consumer.

*C. Reinforcement learning*

The RL deals with knowledge acquisition on how agents should figure out the sequences of actions in a given environment in order to maximize cumulative reward. To identify the best policy for a given goal, the RL algorithms use a variety of ways to optimize a policy [56]. The SARSA [57], policy gradient, and Q-learning are examples of RL algorithms, and these techniques update the policy utilizing a function approximator to track down the ideal arrangement [58]. Moreover, it is feasible to successfully gain optimal policy from a vast measure of information, and this has the benefit of further developing the learning execution by applying different DL strategies.

Convolution in GCNs is basically used to multiply the input of neurons with a set of weights that are commonly known as filters or kernels. Within the same layer, the same filter to the entire image is used, and is referred to as weight sharing. Similar actions are carried out by GCNs; however, the model learns the features by looking at the nearby nodes. The primary distinction between CNNs and GCNs is that the former was developed specifically to function on regular (Euclidean) organized data, whilst the latter were generalized CNNs with varying numbers of connections and unordered nodes. Spatial and spectral GCNs are generally used for time series image data [59].

*1) Deep Q-learning algorithm*

RL has grown in popularity in recent years due to its efficacy in addressing difficult sequential decision-making situations, and some of these accomplishments can be attributed to the use of RL [60]. The DQN algorithm, which was introduced in [60], combines the traditional Q-Learning with neural networks [61, 62].

Using the identical architecture, method and hyperparameters for each game, the authors demonstrated that their algorithm can play Atari 2600 games at a professional human level [60, 61]. The DQN has a strong function approximation ability that allows action value functions to converge faster and with fewer training iterations [63]. It is a derivative of conventional Q-Learning and approaches the ideal action-value function given below:

$$Q(s,a) = \gamma max(Q(s',a')) + r(s,a) \tag{1}$$

where $Q(s,a)$ is the Q function of taking action $a'$ in state $s'$, $\gamma$ is the discount factor having real value $\in [0,1)$ which cares for the reward agent achieved in the past, present, and future, $r(s,a)$ is the reward function of taking action $a$ in state $s$. The $max(Q(s',a'))$ is the maximum possible Q value of the next state $s'$ getting the best action $a'$. The DQN performs a Q-Learning update at every time step that uses the loss function defined as follows:

$$L(\theta) = \left[\left(r(s,a) + \gamma \, max(Q(s',a';\theta))\right) - Q(s,a;\theta)\right]^2 \tag{2}$$

where $L(\theta)$ is the loss function taking parameter $\theta$, $\left(r(s,a) + \gamma \, max(Q(s',a';\theta))\right)$ is the target Q function, $Q(s,a;\theta)$ is the predicted Q function. The network of RL is trained by minimizing $L(\theta)$ compared with the target and predicted Q function, and finally $Q(s,a;\theta)$ of RL is estimated.

*2) Deep Recurrent Q-learning algorithm*

A deep recurrent Q-network (DRQN), which is a combination of a deep Q-network and a recurrent LSTM layer, is introduced to integrate time sensitive and sequential information with standard Atari games [64]. The DQN is modified to the DRQN by replacing the first convolutional layer with the LSTM layer in the DQN model. The LSTM layer of the DRQN provides a memory function, which allows new input data to influence the Q-values of the previous data. Generally, LSTM cells have been employed for various recurrent neural network (RNN) applications such as video classification [65] and language translation [66]. In deep RL models with a recurrent structure, the DRQN-based agents can learn the enhanced policy with a robust sense under the partially observable environment. Unlike the DQN, the Q-function of the DRQN is approximated as $Q(o,a)$ which obtains an observation $o$ and an action $a$ by the RNN. The LSTM cell consists of an input gate, a forget gate, and an output gate. The data, which is maintained throughout the entire sequence being input to the RNN, is included in the cell. Moreover, the cell that serves as memory for the RNN is controlled by the different gates of the LSTM node to contain significant information or lose unimportant information. The input gate updates the cell with fresh data, and the forget gate removes inconsistent data from the cell vector. The output gate selects the aspect of the fresh data that is sent into the RNN and changed by the cell.

*3) Bi-directional Deep Recurrent Q-learning algorithm*

The LSTM network is a particular kind of deep neural network proposed by Hochreiter and Schmidhuber [67]. It is challenging to train deep neural network with time series data, due to its memory-bandwidth-bound processing requirements. To overcome the challenges of the LSTM network and achieve more effective prediction of engineered systems under complicated operational settings, Bi-LSTM network was utilized, which is described as a sequence processing model [68]. Two principal hidden layers make up the overall structure of the Bi-LSTM network. A bidirectional memory is added to the network by stacking identical LSTM networks in each layer, with one serving as a backward hidden layer and the other as a forward hidden layer. They provide a bi-directional memory function for the network. Distinct from the LSTM, the Bi-LSTM obtains the future information while implementing it in the backward direction, which stores the past and future information by adopting the two LSTM cells in opposite directions [69]. In fact, the Bi-LSTM has a robust memory to accurately preserve all of the useful past and future features [70]. The Bi-LSTM has higher precision to predict phenomena with high stochastic and irregular behavior as it does not follow the recursive process, which causes error accumulation by reflecting the past information in an iterative way. In addition, the Bi-LSTM improves the performance of the algorithm by increasing the amount of information used in the network.

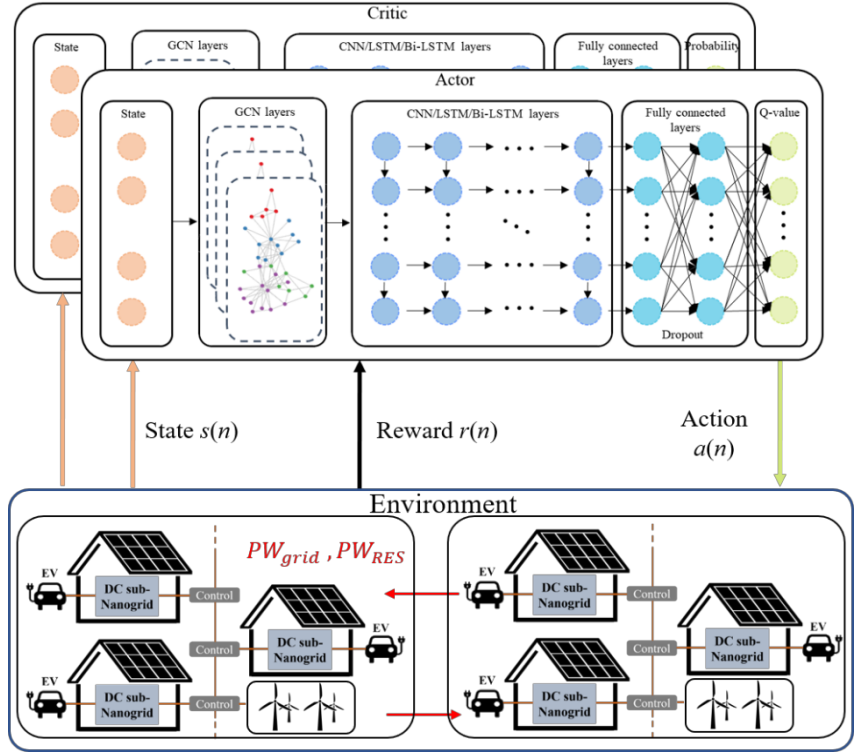

Fig. 3. Architecture of GCN-DQN, GCN-DRQN, GCN-Bi-DRQN applied to the P2P power trading system.

*4) Proximal policy optimization (PPO)*

Trust region policy optimization (TRPO) is a prevalent approach in RL [71]. TRPO's restriction of the Kullback-Leibler (KL) divergence between succeeding policies on the optimization trajectory is given as:

$$maximize\, \mathbb{E}\left[r_t(\theta)\hat{A}_t\right] = \mathbb{E}\left[\frac{\pi_\theta(a_t|s_t)}{\pi_{\theta_{old}}(a_t|s_t)}\hat{A}_t\right] \tag{3}$$

$$\text{subject to } \mathbb{E}\left[D_{KL}(\pi_\theta(\cdot|s_t))||(\pi_{\theta_{old}}(\cdot|s_t))\right] \le \delta, \forall s \tag{4}$$

where $r_t(\theta)$ is the reward, $\hat{A}_t$ is the advantage, $a_t$ is the action, $s_t$ is the state, $\pi_\theta$ and $\pi_{\theta_{old}}$ are policies, $\mathbb{E}$ is the expected value of the loss function, $D_{KL}$ is the KL divergence, and $\delta$ is bound on the KL divergence.

The PPO is a well-known deep RL algorithm that has demonstrated success on challenging tasks like playing video games and manipulating dexterous robots [72, 73]. It is an actor-critic algorithm that retrieves a policy (the actor) and a value-function (the critic) and calculates the value of the discounted reward that the agent would receive at the conclusion of the process starting from any state. The policy loss function used for PPO is given below:

$$L^C(t) = \mathbb{E}\left[min(\underbrace{r_t(\theta)\hat{A}_t}_{surrogate\,objective}, clip(r_t(\theta), 1-\epsilon, 1+\epsilon)\hat{A}_t)\right] = \mathbb{E}\left[min(\underbrace{\frac{\pi_\theta(a_t|s_t)}{\pi_{\theta_{old}}(a_t|s_t)}\hat{A}_t}_{surrogate\,objective}, clip(\frac{\pi_\theta(a_t|s_t)}{\pi_{\theta_{old}}(a_t|s_t)}, 1-\epsilon, 1+\epsilon)\hat{A}_t)\right]$$

(5)

where $L^C(t)$ is the loss function with a clipped surrogate objective function to constrain proposed policy update with a new objective function, $\epsilon$ is a hyperparameter. Since we initialize $\pi_\theta$ as $\pi_{\theta_{old}}$ (and thus the ratios all start from one), the first step we take is identical to a maximization step over the unclipped surrogate reward. Therefore, the size of the surrogate landscape alone determines how big a step we take, and we can end up traveling arbitrarily far from the trust area.

### III. P2P POWER TRADING

#### A. RL-based P2P power trading

*1) P2P power trading strategy*

A cooperative game model for P2P power trading is applied to real-time power management of the nanogrid cluster. The cooperative game model is used so that independent clusters work together as a single entity to minimize electricity costs through P2P power trading. In P2P power trading, buying and selling rules between clusters are set according to two cases. Case 1 corresponds to a situation where the amount of RES power supplied by the producer is greater than the amount of power required by the consumer, and case 2 refers to a situation where the amount of RE power supplied by the producer is less than the amount of power required by the consumer. P2P power trading between nanogrid clusters is implemented individually using UT and RESs. It can take advantage of other households by buying their excess power at a relatively cheaper price with respect to high price of the grid power. Each cluster has 3 and 5 modes of P2P power trading for economical operation and peak demand benefits.

*2) Model of action space*

The P2P power trading process is established as a Markov decision process (MDP) with multiple variables, and uses RL technology to achieve the economical operation of nanogrid clusters with cooperative P2P power trading. In an MDP, an action can be selected among the set of all actions by the agent in a given state. In P2P power trading, the set of actions that a prosumer can select in a given state represents the actions for participating in the trading, such as buying, selling, and holding in connection with UT and RESs. The prosumer can choose to buy or sell energy considering the energy deficit or surplus conditions in cooperative P2P power trading. Two types of RL-based P2P power trading methods are presented. P2P power trading with *DQN, DRQN, Bi_DRQN, PPO* considering only RESs defined in equation (6) and the proposed P2P power trading with *N_DQN, N_DRQN, N_Bi_DRQN, N_PPO* considering both UT and RESs presented in equation (7). In equation (6), $A_{P2P}(n)$ and $PW_{P2P}(n)$ are switching functions of action and the amount of traded power in P2P power trading at the *n*-th time interval, respectively. This can be calculated according to the allowable power consumption constraint ($PW^{max}$). In our proposed method, we represent $A_{P2P}(n)$ by $A_{N\_P2P}(n)$ and $PW_{P2P}(n)$ by $PW_{N\_P2P}(n)$, and are defined in equation (7).

$$A_{P2P}(n), PW_{P2P}(n) = \begin{cases} +1(BuyfromRES), & max\{0, (PW_{cluster}(n) - PW_{RES}(n)) - PW^{max}\} \\ -1(SelltoRES), & max\{0, PW^{max} - (PW_{cluster}(n) + PW_{RES}(n))\} \\ 0(idle), & 0 \end{cases}$$

(6)

$$A_{N\_P2P}(n), PW_{N\_P2P}(n) = \begin{cases} +2(BuyfromUT), & max\{0, (PW_{cluster}(n) - PW_{RES}(n)) - PW^{max}\} \\ +1(BuyfromRES), & max\{0, (PW_{cluster}(n) - PW_{RES}(n)) - PW^{max}\} \\ -2(SelltoUT), & max\{0, PW^{max} - (PW_{cluster}(n) + PW_{RES}(n))\} \\ -1(SelltoRES), & max\{0, PW^{max} - (PW_{cluster}(n) + PW_{RES}(n))\} \\ 0(idle), & 0 \end{cases}$$

(7)

where $PW^{max}$ is the allowable power consumption, $PW_{cluster}(n)$ is the required power demand of a nanogrid cluster, and $PW_{RES}(n)$ is the produced power from RESs located in the nanogrid cluster.

In equation (6), we define the P2P power trading actions and traded power with RES by three possible options: +1 (buy from RES, $max\{0, (PW_{cluster}(n) - PW_{RES}(n)) - PW^{max}\}$), -1 (sell to RES, $max\{0, PW^{max} - (PW_{cluster}(n) + PW_{RES}(n))\}$), and 0 (idle,0). In equation (7), we define the proposed power trading actions with UT and RES by five possible options: +2 (buy from UT, $max\{0, (PW_{cluster}(n) - PW_{RES}(n)) - PW^{max}\}$), +1 (buy from RES, $max\{0, (PW_{cluster}(n) - PW_{RES}(n)) - PW^{max}\}$), -2 (sell to UT, $max\{0, PW^{max} - (PW_{cluster}(n) + PW_{RES}(n))\}$), -1 (sell to RES, $max\{0, PW^{max} - (PW_{cluster}(n) + PW_{RES}(n))\}$), and 0 (idle, 0). To maximize the profit of nanogrid clusters through P2P power trading, the UT and RESs are considered in the proposed method.

*3) Model of state space*

The set of states $S_{P2P}(n)$ of the agent, which is the decision maker in a trading environment for *DQN, DRQN, Bi_DRQN, PPO, N_DQN, N_DRQN, N_Bi_DRQN, N_PPO*, is described in equation (8).

$$S_{P2P}(n) = [D_1(n), \dots, D_{10}(n), S_1(n), \dots, S_{10}(n), DR(n), SMP(n)]^T$$

(8)

where $D_i(n)$ is the demand of each cluster, $S_i(n)$ is the supply from renewable power generation, and $DR(n)$ and $SMP(n)$ are represented by DR and SMP, respectively. In the P2P power trading system, $S_{P2P}(n)$ represents the state of the prosumer considering the information of distributed energy production, consumption, and reserves.

*4) Model of reward function*

The reward $R_{P2P}(n)$ that the agent obtains from each action environment for *DQN, DRQN, Bi_DRQN, PPO, N_DQN, N_DRQN, N_Bi_DRQN, N_PPO* in a given state is presented in equation (9). It is typically a scalar, and depends on how the condition is set in each state, the reward obtained by the agent varies. There are two types of reward: immediate reward, which is reward for the outcome of the next state, and delayed reward, which is reward for future results that are affected by the current behavior [27]. In the proposed P2P power trading system, a different immediate reward is given by comparing the electricity cost of a local nanogrid cluster using the baseline method of P2P power trading with the proposed method of P2P power trading using RL technology. A positive reward is provided when the electricity cost of the proposed method is lower than that of the baseline method, and a negative reward is provided when the electricity cost of the proposed method is higher than that of the baseline method.

$$R_{P2P}(n) = \begin{cases} +1 & If(C_{baseline}(n) > C_{proposed}(n)) \\ -1 & If(C_{baseline}(n) \leq C_{proposed}(n)) \end{cases} \tag{9}$$

$$C_{baseline}(n), C_{proposed}(n) = \begin{pmatrix} PW_{grid,load}(n) * O_{grid,load}(n) * DR(n) + PW_{RES,load}(n) * O_{RES,load}(n) * SMP(n) \\ +PW_{grid,P2P}(n) * A_{grid,P2P}(n) * DR(n) + PW_{RES,P2P}(n) * A_{RES,P2P}(n) * SMP(n) \end{pmatrix} \tag{10}$$

where $C_{baseline}(n)$ and $C_{proposed}(n)$ represent the electricity cost of the baseline and the proposed P2P power trading method environment for *DQN, DRQN, Bi_DRQN, PPO, N_DQN, N_DRQN, N_Bi_DRQN, N_PPO*, $PW_{grid,load}(n)$ represents the load supplied by UT, $PW_{RES,load}(n)$ represents the load supplied by RESs, $PW_{grid,P2P}(n)$ is the load for P2P power trading provided by UT, $PW_{RES,P2P}(n)$ is the load for P2P power trading provided by RESs, $A_{grid,P2P}(n)$ is the action state of P2P power trading with UT, $A_{RES,P2P}(n)$ is the action state of P2P power trading with RESs, $O_{grid,load}(n)$ is the switching function of the load with UT, and $O_{RES,load}(n)$ is the switching function of the load with RESs.

B. Multi-objective optimization for a local nanogrid

Multi objective optimization suggests the tradeoff solution for power management in nanogrid clusters with P2P trading method for *Baseline, DQN, N_DQN, DRQN, N_DRQN, Bi_DRQN, N_Bi_DRQN, PPO*. It is assumed that a smart meter in a nanogrid cluster can control the operation of all loads, such as electric appliances, EV, and HVAC. In addition, smart meters can monitor the voltage and current signals of an individual nanogrid, and can calculate the power consumption of the nanogrid cluster based on individual nanogrids. For the optimal operation planning of nanogrid clusters, multi-objective optimization can be executed in real time. The objective functions with constraints in multi-objective optimization are used in various combinations, depending on complexity of problems. The ability to switch schedulable loads is achieved through multi-objective optimization, which allows simultaneous attempts to minimize peak loads and grid dependencies.

1) Objective function #1: minimizing the cluster's maximum load

Minimizing the electricity cost by a schedulable load is the first objective function for the power management of nanogrid clusters. This peak load shifting is performed by schedulable loads without considering non-schedulable loads. For residents' convenience, the non-schedulable loads are operated before the schedulable loads. To provide electricity to load demand of a nanogrid cluster, P2P power trading system with power sources (UT, RESs) is used. Through schedulable load scheduling which controls the switching function of schedulable loads, electricity cost of the nanogrid cluster is minimized.

$$\min_{(O_{flex,1},\dots,O_{flex,I},O_{EV,k})} \begin{bmatrix} \sum_{i=1}^{I} PW_{flex,i}(n) * O_{flex,i}(n) \\ + \sum_{j=1}^{J} PW_{non-flex,j}(n) * O_{non-flex,j}(n) \\ + \sum_{k=1}^{K} PW_{EV,k}(n) * O_{EV,k}(n) \end{bmatrix} \tag{11}$$

subject to

$$SOC_{EV,min}(n) < SOC_{EV,l}(n) < SOC_{EV,max}(n) \ \ l = 1, \dots, L$$

$$\sum_{m=1}^{M} PW_{flex,m}(n) * O_{flex,m}(n) + \sum_{p=1}^{P} PW_{non-flex,p}(n) * O_{non-flex,p}(n) + \sum_{q=1}^{Q} PW_{EV,q}(n) * O_{EV,q}(n) < PW^{max}$$

$$d_{EV,r}(n) < d_{EV,max}(n) \ \ r = 1, \dots, R$$

$$d_{flex,s}(n) < d_{flex,max}(n) \ \ s = 1, \dots, S$$

where $PW_{flex,i}(n)$, $PW_{non-flex,j}(n)$, and $PW_{EV,k}(n)$ are the power consumption of a flexible load, a non-flexible load, and an EV charging/discharging load at the *n*-th time interval, respectively. $O_{flex,i}(n)$, $O_{non-flex,j}(n)$, and $O_{EV,k}(n)$ are the switch functions of a flexible load, a non-flexible load, and EV charging/discharging in the nanogrid cluster at the *n*-th time interval, respectively. The increment of *n* by 1 at the *n*-th time interval corresponds to an additional 10 min time interval. The $SOC_{EV,l}(n)$ is the SOC of the *l*-th EV load, $SOC_{EV,min}(n)$ and $SOC_{EV,max}(n)$ represent the minimum and maximum SOC of EV charging/discharging at n-th time interval, $d_{EV,r}(n)$ and $d_{flex,s}(n)$ represent the accumulated delay in scheduling the EV and flexible load, $d_{EV,max}(n)$ and $d_{flex,max}(n)$ represent the maximum allowable accumulated delay of EV and flexible load, respectively. The accumulated delay in scheduling the EV and flexible load is evaluated according to the number of requests using the EV and flexible load.

*2)Objective function #2: maximizing the cluster's RESs' dependence for carbon neutrality*

For the eco-friendly operation and carbon neutrality of nanogrid clusters, the ability to supply power from RESs is becoming more crucial. Although it is more economical to consume power from the UT at a lower rate depending on the DR program over time, prioritizing the use of PV power is promoted for environmental operation management. As more power is consumed locally, instead of passing through long transmission lines without power loss, RESs have a significant impact on fossil energy reduction and energy efficiency improvement.

$$\min_{\begin{pmatrix} PW_{grid,load}, O_{grid,load} \\ ,PW_{RES,load}, O_{RES,load} \\ ,PW_{EV,load}, O_{EV,load} \end{pmatrix}} \begin{bmatrix} PW_{grid,load}(n) * O_{grid,load}(n) - PW_{grid,P2P}(n) * A_{grid,P2P}(n) \\ -PW_{RES,load}(n) * O_{RES,load}(n) - PW_{RES,P2P}(n) * A_{RES,P2P}(n) \\ -PW_{EV,load}(n) * O_{EV,load}(n) - PW_{EV,P2P}(n) * A_{EV,P2P}(n) \end{bmatrix} \tag{12}$$

subject to

$$PW_{grid,load}(n) + PW_{grid,P2P}(n) < PW_{grid,max}(n)$$

$$PW_{RES,load}(n) + PW_{RES,P2P}(n) < PW_{RES}(n)$$

Self-supplied power from RESs is preferentially used to reduce reliance on the UT, so that the RESs can provide part of the power consumption for load demand and the rest can be supplied from the UT. P2P trading for the baseline model is implemented using multi-objective optimization defined by the equations (11) and (12), considering the constraints $PW_{grid,load}(n) + PW_{grid,P2P}(n) < PW_{grid,max}(n)$ and $PW_{RES,load}(n) + PW_{RES,P2P}(n) < PW_{RES}(n)$. The *DQN, DRQN, Bi_DRQN, PPO* methods of RL-based cooperative P2P power trading in equations (6), (8), (9), and (10) are used with multi-objective functions (11) and (12). *N_DQN, N_DRQN, N_Bi_DRQN, N_PPO* methods of the proposed RL-based cooperative P2P power trading in equations (7), (8), (9), and (10) are used with multi-objective functions (11) and (12). With the RL-based cooperative P2P trading, prosumers and consumers can use $A_{P2P}(n), PW_{P2P}(n), A_{N\_P2P}(n), PW_{N\_P2P}(n)$ in equations (6) and (7) within the constraints.

IV. CASE STUDY

*A. Experiment setup*

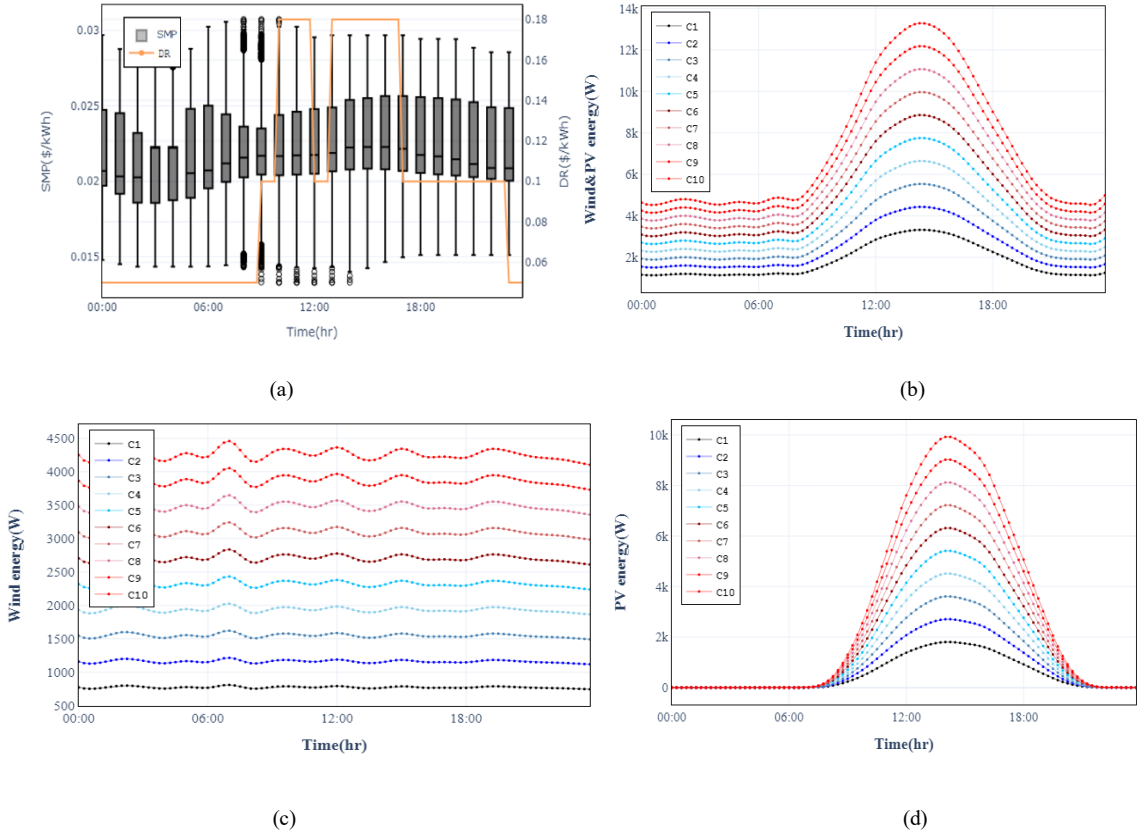

Fig. 4. Variations of electricity rate and renewable energy production by clusters: (a) Average electricity rate of DR program (ToU tariff) and SMP on yearly basis; (b) Average Wind & PV energy production by 10 clusters on yearly basis; (c) Average Wind energy production by 10 clusters on yearly basis; (d) Average PV energy production by 10 clusters on yearly basis.

DR programs composed of three types of electricity rates, namely, the ToU-based electricity rate, the

progressive-based electricity rate, and the climate change & environmental charge rate are defined as:

$$DR = (R_{TOU}, R_{PROG}, R_{CCEC}) \tag{13}$$

$$R_{TOU} = \begin{cases} 0.06\$/kWh(off-peak) & If(23:00 < t \leq 09:00) \\ 0.1\$/kWh(mid-peak) & If \begin{pmatrix} 09:00 < t \leq 10:00 \\ or\,12:00 < t \leq 13:00 \\ or\,17:00 < t \leq 23:00 \end{pmatrix} \\ 0.18\$/kWh(on-peak) & If \begin{pmatrix} 10:00 < t \leq 12:00 \\ or\,13:00 < t \leq 17:00 \end{pmatrix} \end{cases} \tag{14}$$

$$R_{PROG} = \begin{cases} 0.008\$/kWh & If(PW_{total} \leq 300kWh) \\ 0.018\$/kWh & If(301kWh < PW_{total} \leq 450kWh) \\ 0.027\$/kWh & If(450kWh < PW_{total}) \end{cases} \tag{15}$$

$$R_{CCEC} = R_{RPS} + R_{ETS} + R_{CGR} \tag{16}$$

where $R_{TOU}$ is the ToU-based electricity rate, $R_{PROG}$ is the progressive-based electricity rate, $R_{CCEC}$ is the climate change & environmental charge rate related to carbon neutrality, $R_{RPS}$ is the renewable portfolio standards (RPS) electricity rate, $R_{ETS}$ is the emission trading system electricity rate, and $R_{CGR}$ is the coal generation reduction (CGR) electricity rate. The $PW_{total}$ is the total power consumption in a month, which is calculated by the summation of the power consumption of schedulable and non-schedulable loads. The $R_{TOU}$ varies according to the off-peak, mid-peak, and on-peak time intervals mentioned in equation (14). The $R_{PROG}$ is determined according to $PW_{total}$, defined in equation (15), and the $R_{CCEC}$ is calculated by taking the sum of $R_{RPS}$, $R_{ETS}$, and $R_{CGR}$ defined in equation (16).

The electricity cost is calculated with the DR program and the SMP as shown in Fig. 4(a). In Fig. 4(a), yellow line represents the DR program plotted across the right vertical axis, and the SMP is plotted across the left vertical axis. Moreover, Fig. 4(a) shows the boxplot of SMP, and is helpful to see the differences between data points. They are able to deliver a substantial information of statistical data, including medians, ranges, and outliers. Line in between the rectangles are whiskers differentiating scores between lower and upper quartile, and the black circles outside the whiskers of box plot represent the outlier. Due to the presence of outliers, fluctuations occur in SMP. Simulations are implemented with renewable energy production by each cluster shown in Figs. 4(b-d). The electricity cost of the cluster increases according to the intermittency of renewable energy. In the hybrid wind PV system, it confirms that the effect of P2P power trading is huge as it is well distributed in the situation of insufficient power. Cluster 10 produces significantly larger amount of renewable energy than cluster 1, as presented in Fig. 4(b).

The simulations are implemented with P2P power trading for *Baseline* (rule-based model), *DQN, N_DQN, DRQN, N_DRQN, Bi_DRQN, N_Bi_DRQN, PPO* and *N_PPO*. Each layer of RL is based on the GCN-CNN layers (*DQN, N_DQN*), GCN-LSTM layers (*DRQN, N_DRQN*) and GCN-Bi-LSTM layers (*Bi_DRQN, N_Bi_DRQN, PPO, N_PPO*). The efficiency of our proposed technique is shown by comparing it to the following techniques:

1) *Baseline* (rule-based model): The clusters exclusively sell renewable energy with external networks and do

not employ RL for P2P power trading. The rule-based operating policy calculates the difference between the estimated energy demand and generation for the trading time period, and a cluster sells the surplus electricity or buys the needed energy from other clusters according to the situation.

2) *DQN, DRQN, Bi_DRQN, PPO*: The *DQN, DRQN, Bi_DRQN, PPO* methods are configured similarly to the rule-based one, with the exception that it uses ten independent cluster agents to track each cluster's trading activity on external networks. As a result, the agents train their critic networks using their own observations, actions, reward, and subsequent observations.

3) *N_DQN, N_DRQN, N_Bi_DRQN, N_PPO*: The *N_DQN, N_DRQN, N_Bi_DRQN, N_PPO* methods use the proposed model for P2P power trading with UT and RESs. In the P2P power trading, consumers and prosumers can opt for the reduction of electricity costs and the intermittency of self-supplied power generation by strategically using the UT and RESs. The electricity rate of the UT and self-distributed power generation varies over time, and it becomes cheaper as more electricity is used appropriately every hour. A hybrid renewable energy system consisting of wind and PV systems can increase the efficiency of P2P trading and bring more profits to consumers and prosumers considering the electricity costs that change over time.

Figure 5 shows the average reward for the training epoch with *DQN, DRQN, Bi_DRQN, PPO*, and proposed DQN, DRQN, Bi_DRQN, PPO techniques represented by N_DQN, *N_DRQN, N_Bi_DRQN* and *N_PPO*. The hyperparameters used for simulations are adopted from [45, 73, 74] describing the environment of the *DQN, N_DQN, DRQN, N_DRQN, Bi_DRQN, N_Bi_DRQN, PPO* and *N_PPO*. The hyperparameters for *DQN, N_DQN, DRQN, N_DRQN, Bi-DRQN,* and *N_Bi_DRQN* are: optimizer=Adam, batch size $B$ =128, $N_T$=22, $\gamma$=0.95, $\epsilon$=0.1, $\epsilon_{decay}$=0.995, $\epsilon_{min}$=0.01, learning rate=0.005. The hyperparameters for *PPO, N_PPO* setup are: $\gamma$=0.99, update interval=128, actor learning rate=0.0005, critic learning rate=0.001, clip ratio=0.1, $\lambda$ =0.95, epochs=3. To prove the effectiveness of the proposed P2P trading, we present a comparison of the training performance between the proposed P2P trading and the typical P2P trading methods. The comparison makes it clear that the *N_DQN, N_DRQN, N_Bi_DRQN,* and, *N_PPO* methods outperform the *DQN, DRQN, Bi_DRQN,* and *PPO* methods by getting more reward (accumulated trading profits) as shown in Fig. 5.

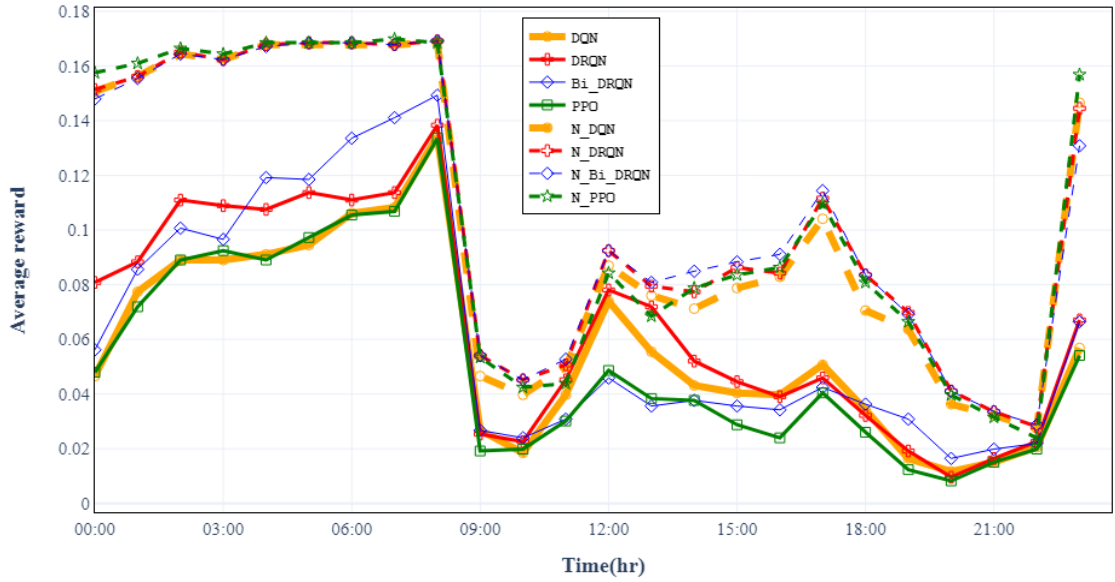

Fig. 5. Average reward for each training epoch with DQN, DRQN, Bi_DRQN, PPO, N_DQN, N_DRQN, N_Bi_DRQN and, N_PPO techniques.

*B. Case study of experimental results*

Figure 6 shows the total cluster electricity cost of schedulable/non-schedulable electric appliances in a nanogrid cluster corresponding to clusters 1, 5, and 10. In the *Baseline*, *DQN*, *DRQN*, *Bi_DRQN,* and *PPO* cases*,* the selling/buying operation of each cluster is determined, considering only the SMP, and in the *N_DQN, N_DRQN, N_Bi_DRQN, N_PPO* cases, the selling/buying operation of each cluster is determined according to real-time fluctuations of the DR related to carbon neutrality and SMP, which achieves more flexible power management. In Fig.5, reward is obtained during the training in all the off-peak, mid-peak, and on-peak time intervals of DR. The intermittent nature of PV energy is compensated by wind energy produced at every peak period, and the proposed P2P trading system works well with it. As a result, there is reduction of electricity costs in all intervals shown in Fig. 6. Even in areas with low renewable energy production, it can be confirmed that electricity costs are lowered through RL-based cooperative P2P power trading using DR and SMP. The proposed EV charging/discharging strategy in conjunction with P2P power trading provides electricity cost reduction depending on the trend of wind and PV power generation in real time. The proposed approach can reduce the electricity cost and outperform other methods. The *N_PPO* approach achieves an average electricity cost of $641.5, $602.5 and $193.2 in clusters 1, 5, and 10, respectively. In relative terms, *N_PPO* reduces the average electricity cost by 5.99%, 27.00%, and 74.78% compared with the *Baseline* for clusters 1, 5, and 10 as shown in Table 2, respectively. Compared to the case where the amount of renewable energy generation in cluster 1 is small, the proposed algorithm shows better performance as the renewable energy generation in cluster 10 is large as shown in Fig. 6.

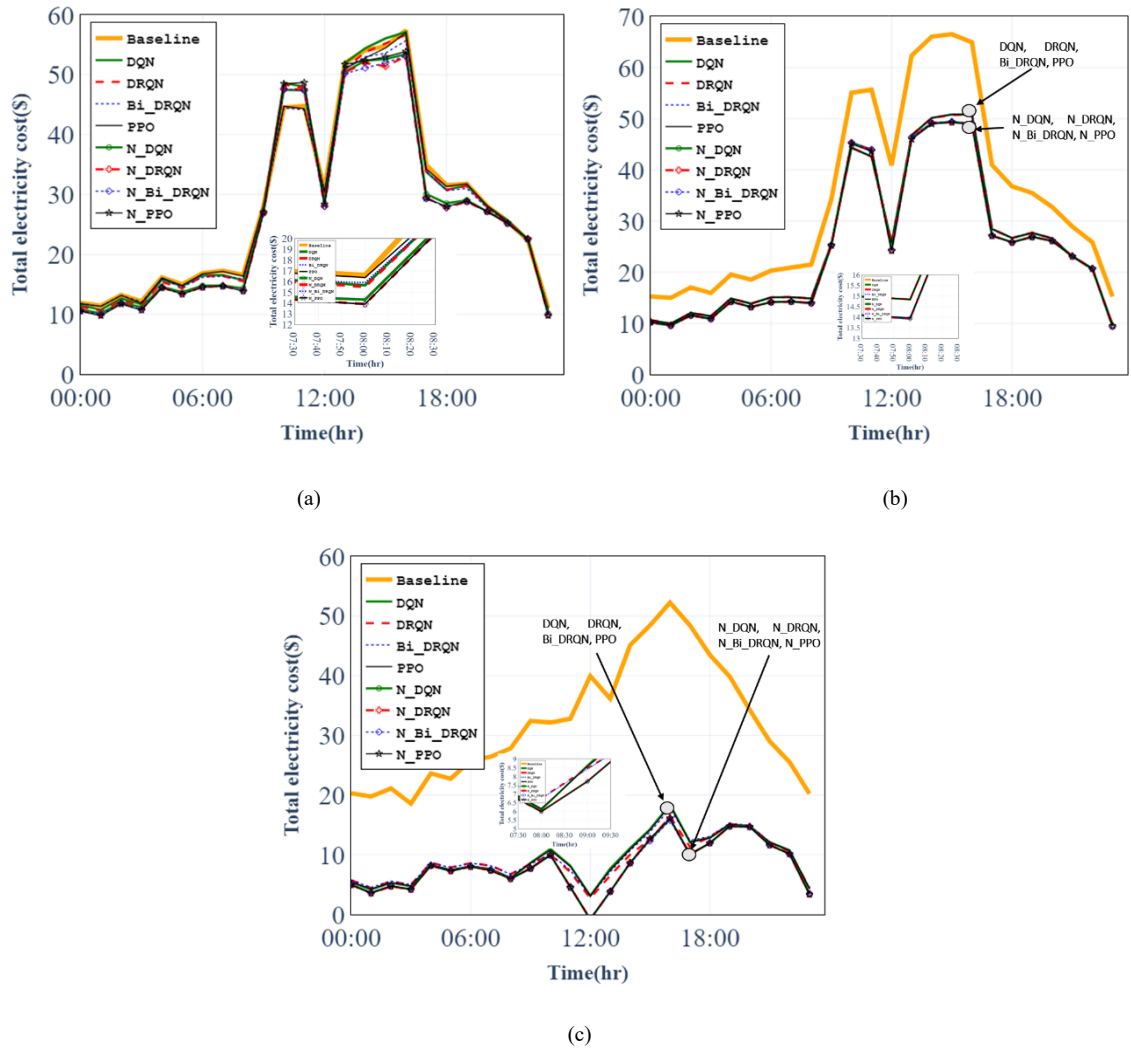

Fig. 6. Average electricity cost of single cluster on yearly basis with schedulable/non-schedulable electric appliances: (a) cluster 1; (b) cluster 5; (c) cluster 10.

The total cluster power consumption for schedulable/non-schedulable electric appliances of the nanogrid clusters is depicted in Fig. 7. Unlike electricity costs, the total cluster power consumption (sum of flex, non-flex, and EV power) of the *N_DQN, N_DRQN, N_Bi_DRQN, N_PPO* has a higher value compared with the *Baseline*, *DQN*, *DRQN*, *Bi_DRQN*, *PPO* in clusters 1, 5, and 10. By reducing the scheduling delay and effectively using additional renewable energy with the P2P power trading system, the total cluster electricity cost can be decreased although the total cluster power consumption increases. In all peak intervals of DR, with a low renewable energy production rate from cluster 1 to high renewable energy production rate in cluster 10, electricity costs were minimized through RL-based cooperative P2P power trading with EV charging/discharging. Th objective functions #1, #2 that meet the goal of carbon neutrality for local power management led to the reduction of total power consumption.

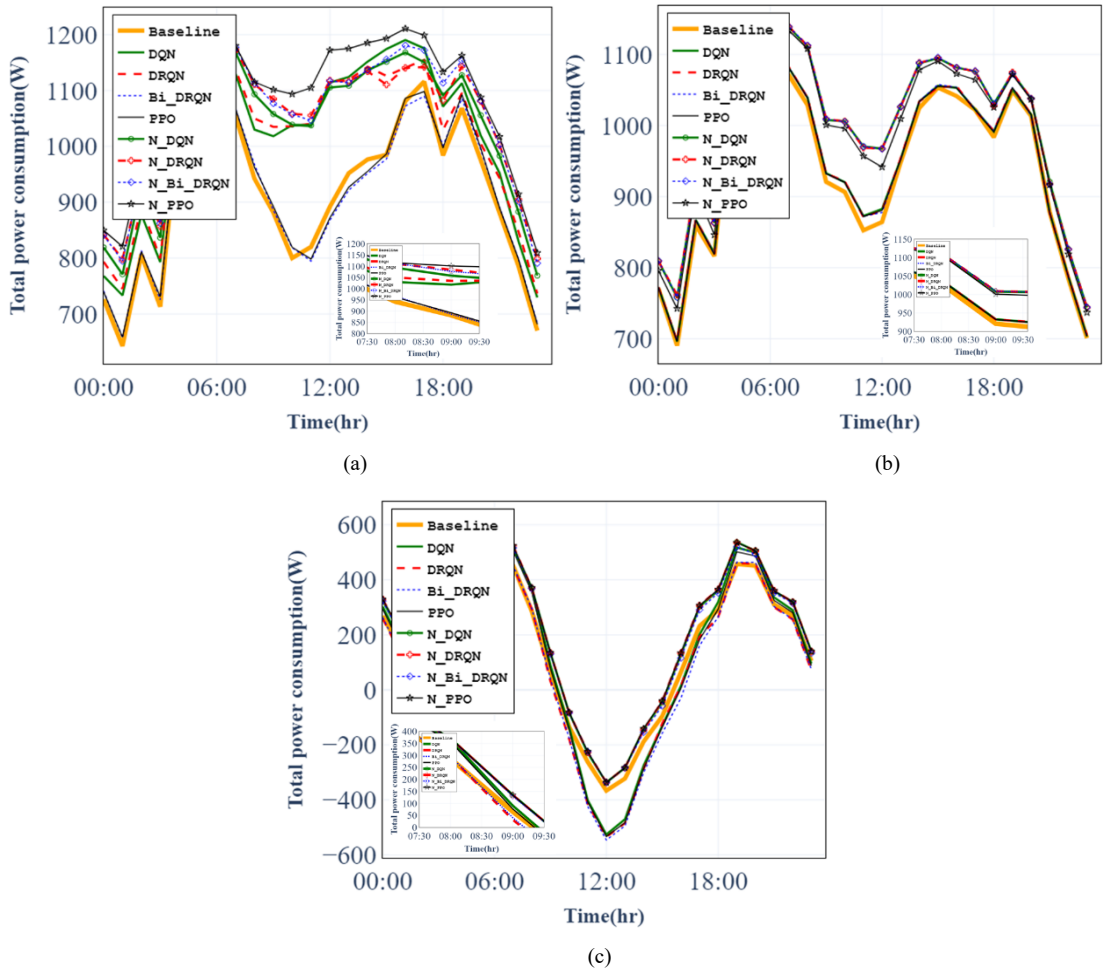

Fig. 7. Average power consumption of single cluster on yearly basis with schedulable/non-schedulable electric appliances: (a) cluster 1; (b) cluster 5; (c) cluster 10.

Figure 8 shows the average traded power of nanogrid clusters by discharging EV and utilizing renewable energy systems for clusters 1, 5, and 10. As shown in Fig. 8, the traded power of the *DQN, DRQN, Bi_DRQN, PPO, N_DQN, N_DRQN, N_Bi_DRQN,* and *N_PPO* to minimize scheduling delay through P2P power trading is higher than that of rule-based methods. In Fig. 8(a), the renewable energy production of cluster 1 is small, so it is used for power consumption without P2P power trading. The reason for the high electricity cost even with a large traded power is that DR&SMP are not well considered in cluster 1. Since clusters 5 and 10 have the higher capacity of wind and PV energy production, more P2P power trading is performed in clusters 5 and 10 compared with the cluster 1. The flexibility of each cluster is improved, and the intermittent part of the renewable energy is mitigated. The results also confirm that the power from EV discharging and renewable energy systems is well distributed to each nanogrid cluster where energy is insufficient through P2P power trading.

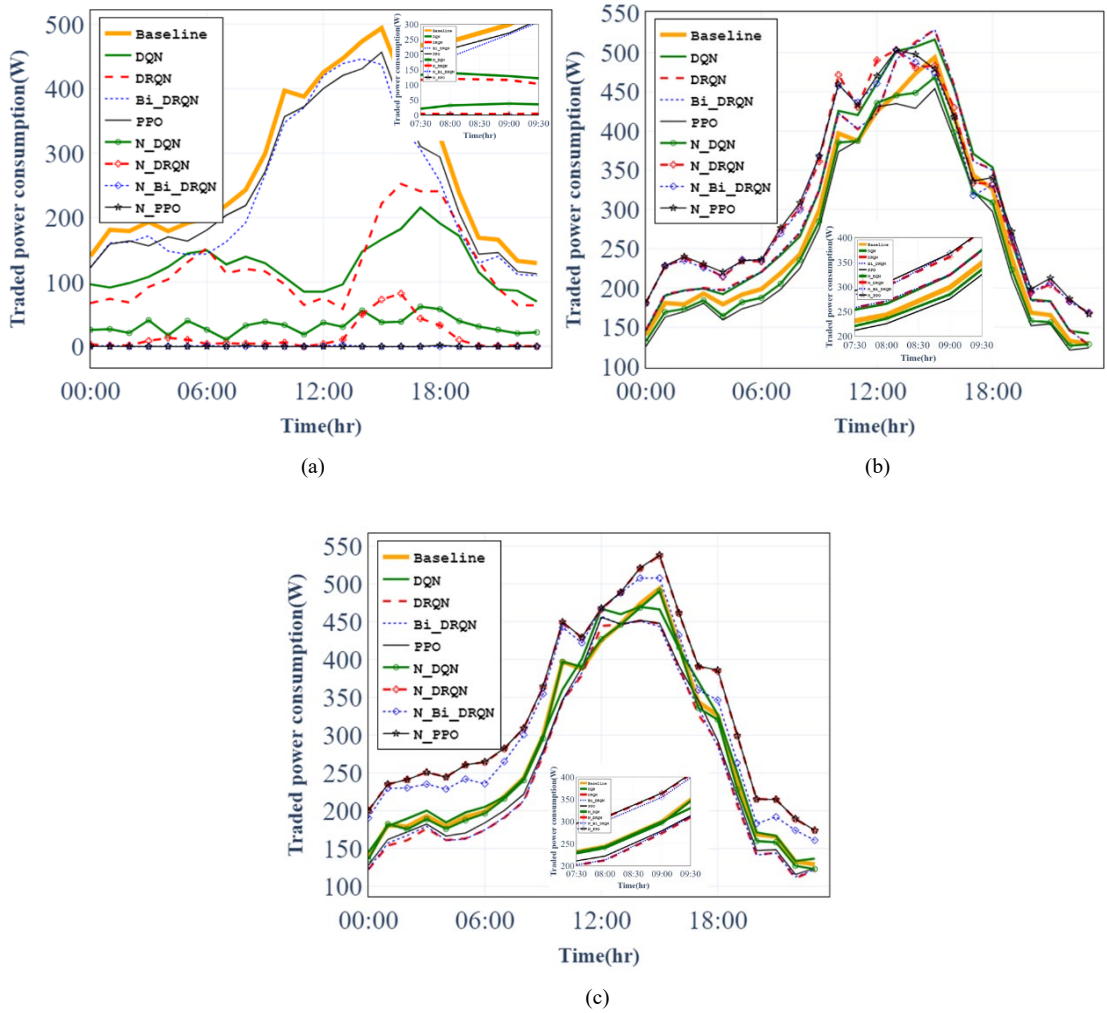

Fig. 8. Average traded power of single cluster on yearly basis by discharging EV and utilizing renewable energy systems: (a) cluster 1; (b) cluster 5; (c) cluster 10.

TABLE 2 shows the total cluster electricity costs for nine types of techniques. In the proposed method, *N_PPO*, the total average electricity cost reduction of the nanogrid clusters 1, 5 and 10 are 5.99%, 27.00%, and 74.78% and the renewable energy usage increases through power distribution of renewable energy via P2P power trading. With a low renewable energy production rate in cluster 1 and a high renewable energy production rate in cluster 10, the RL-based cooperative P2P power trading can reduce the average electricity cost throughout the DR peak periods. Wind energy produced at every peak interval balances the intermittent nature of the PV power production and thus shows good coordination with the proposed P2P trading method. In Fig.5, it can be seen that the reward of *N_PPO* is the highest and the saving effect is the greatest, which also can be seen in Table 2. Consequently, we can see that RL-based cooperative P2P power trading between EV prosumers and renewable energy generation prosumers reduces electricity costs. The combination of intermittent wind and PV energy generation and P2P power trading shows better results and is recommended as a future power system.

TABLE 2. Average electricity cost and savings of different techniques for clusters 1, 5, and 10.

| Cluster | | Baseline | DQN | DRQN | Bi_DRQN | PPO | N_DQN | N_DRQN | N_Bi_DRQN | N_PPO |
|---|---|---|---|---|---|---|---|---|---|---|
| Cluster 1 | Electricity cost($) | $ 682.4 | $ 682.0 | $ 676.4 | $ 669.6 | $ 676.4 | $ 648.5 | $ 648.2 | $ 642.2 | $ 641.5 |
| | Savings(%) | - | 0.05% | 0.87% | 1.87% | 0.87% | 4.96% | 5.01% | 5.89% | 5.99% |
| Cluster 5 | Electricity cost($) | $ 825.4 | $ 603.3 | $ 617.1 | $ 617.1 | $ 617.2 | $ 603.7 | $ 603.5 | $ 603.7 | $ 602.5 |
| | Saving(%) | - | 26.90% | 25.23% | 25.23% | 25.22% | 26.85% | 26.88%, | 26.85%, | 27.00% |
| Cluster 10 | Electricity cost($) | $ 766.3 | $ 221.5 | $ 215.8 | $ 222.0 | $ 220.6 | $ 193.7 | $ 193.8 | $ 193.6 | $ 193.2 |
| | Saving(%) | - | 71.09% | 71.83% | 71.02% | 71.21% | 74.72% | 74.70% | 74.73% | 74.78% |

The proposed GCN-Bi-LSTM based PPO technique (N_PPO), which obtains reward in all sections through RL-based cooperative P2P power trading shows good performance. Information of adjacent nodes of the GCN and Bi-LSTM network can be collected and the entire graph can be represented as one. Due to the flexibility of the graph structure, not only social networking data that has a graph like structure, but also image data that are expressed in other forms can be given the shape of graph. Although PPO was introduced in 2017, it still shows the best performance in terms of computational efficiency, and is easy to implement.

The above mentioned scenario could be advantageous to the UT and consumers, enabling the effective utilization of demand side resources, the local supply demand balance, as well as opportunities for consumers to gain financial benefits through P2P power trading. The proposed P2P architecture supports the local energy trading and allows users to trade their excess energy with nearby users that help in reducing the electricity costs and the intermittency of energy generation by strategically using the UT and RESs. Next, the impact of RL is further discussed. All the off-peak, mid-peak, and on-peak time periods of DR are given reward during the training, and RL is having an impact on the producer's overall revenues. The cooperative P2P power trading mechanism, developed as a MDP, employs RL technology to enable the economical functioning of nanogrid clusters. The reason for adopting this technique is that an action in an MDP can be chosen by the agent in a particular state from a list of all possible actions. Moreover, in P2P power trading, the set of options that a prosumer can choose from a particular state corresponds to the actions for engaging in trading, such as buying/selling from/to UT and RESs. The selling/buying operation of each cluster in the typical method is determined considering only the SMP, and in the proposed method, this operation is based on the real-time fluctuations of the DR related to carbon neutrality and SMP, which achieves more flexible power management. In simulation results, the proposed technique obtains reward in all sections through RL-based cooperative P2P power trading and applies a model for data handling that shows good performance, and can effectively reduce the investment.

## V. CONCLUSION

We presented a novel cooperative P2P power trading method that adopts reinforcement learning (RL) technique to establish the prosumer model, and applied the method to DC nanogrid clusters for power management through the P2P power trading based on cooperative game theory. In this study, a flexible and reliable DC nanogrid is suitable to be interconnected with a hybrid wind-PV system to utilize RESs, wind and

PV energy, as the secondary energy sources. For maximizing the profit of P2P power trading, not only RESs, but also the utility grid is used in the proposed RL-based cooperative P2P power trading with GCN-Bi-LSTM layers-based PPO technique (*N_PPO*). For the power management of nanogrid cluster, multi-objective optimization is considered to minimize the clusters' maximum load and to maximize the clusters' dependencies on RESs for carbon neutrality. As the P2P power trading method was applied to each simulation environment, the total electricity cost decreased in the aspect of public welfare for buyers and sellers. Moreover, the simulation results revealed the flexibility and reliability of the nanogrid cluster system with P2P power trading. In the proposed method, the P2P power trading and power management in the nanogrid cluster were simulated on the distribution test feeder in real time and the nanogrid cluster was able to reduce the average electricity cost among the RL algorithms used for comparison by 36.7%, averaging over nanogrid clusters.